%% file: bare_jrnl_new_sample4.tex
\documentclass[lettersize,journal]{IEEEtran}
\usepackage{amsmath,amsfonts}
\usepackage{algorithmic}
\usepackage{algorithm}
\usepackage{array}
\usepackage[caption=false,font=normalsize,labelfont=sf,textfont=sf]{subfig}
\usepackage{textcomp}
\usepackage{stfloats}
\usepackage{url}
\usepackage{verbatim}
\usepackage{graphicx}
\usepackage{cite}
\usepackage{booktabs}
\begin{document}

\title{InteractPro: A Unified Framework for Motion-Aware Image Composition}
\author{
    Weijing Tao, 
    Xiaofeng Yang, 
    Miaomiao Cui, 
    Guosheng Lin\thanks{
Weijing Tao, Xiaofeng Yang, and Guosheng Lin are with the College of Computing and Data Science, Nanyang Technological University, Singapore 639798. (e-mail: {weijing002,xiaofeng001}@e.ntu.edu.sg; gslin@ntu.edu.sg)\\
Weijing Tao and Miaomiao Cui are with the DAMO Academy, Alibaba Group, Hangzhou 311121, China. (e-mail: miaomiao.cmm@alibaba-inc.com)
    }
}



\maketitle

\begin{abstract}
We introduce InteractPro, a comprehensive framework for dynamic motion-aware image composition. At its core is InteractPlan, an intelligent planner that leverages a Large Vision Language Model (LVLM) for scenario analysis and object placement, determining the optimal composition strategy to achieve realistic motion effects. Based on each scenario, InteractPlan selects between our two specialized modules: InteractPhys and InteractMotion. InteractPhys employs an enhanced Material Point Method (MPM)-based simulation to produce physically faithful and controllable object-scene interactions, capturing diverse and abstract events that require true physical modeling. InteractMotion, in contrast, is a training-free method based on pretrained video diffusion. Traditional composition approaches suffer from two major limitations: requiring manual planning for object placement and generating static, motionless outputs. By unifying simulation-based and diffusion-based methods under planner guidance, InteractPro overcomes these challenges, ensuring richly motion-aware compositions. Extensive quantitative and qualitative evaluations demonstrate InteractPro’s effectiveness in producing controllable, and coherent compositions across varied scenarios.
\end{abstract}


\begin{figure*}
  \includegraphics[width=\textwidth]{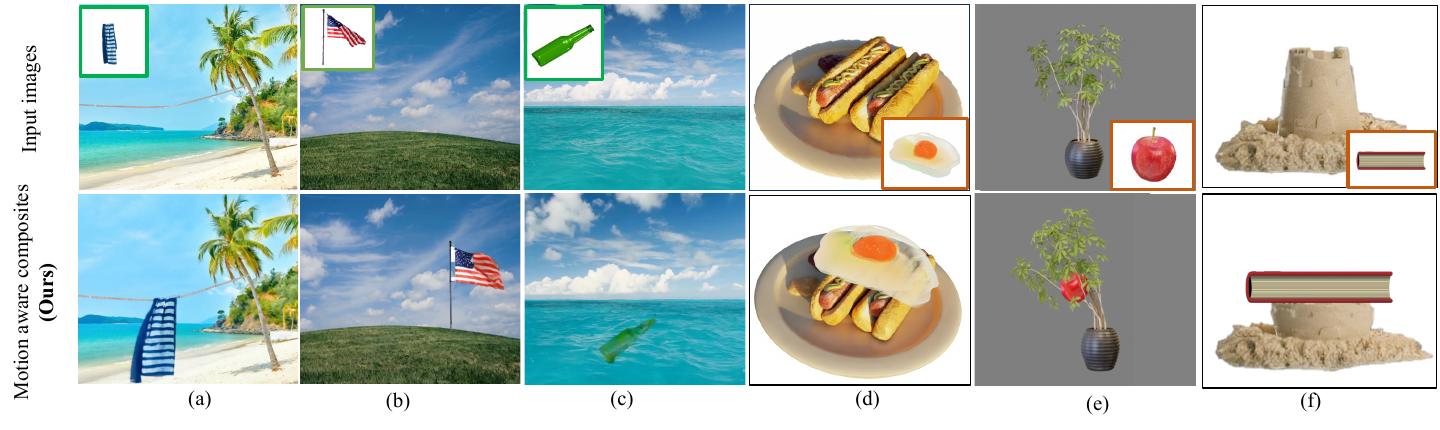} 
  \caption{Introducing InteractPro, a comprehensive framework for motion-aware image composition. InteractMotion (a-c) leverages motion priors from pretrained video diffusion models to generate data-driven dynamics, such as towel and flag fluttering in wind implied by the environment(a-b) and bottle blending seamless into the ocean (c). InteractPhys (d-f) simulates object interactions explicitly through material point method (MPM) physics, capturing effects like runny egg conforming to the shape of hotdog (d), apple wedging between plant branches and causing them to bend (e), and sandcastle compressed under the weight of a book (f). Together, they handle a wide range of scenarios—from learned, appearance-driven motion to physically grounded interactions beyond the scope of diffusion models.}
  \label{fig:teaserall}
\end{figure*}
\section{Introduction}
\IEEEPARstart{H}{ow} can a static object, like a book, be realistically integrated into a background image with a sandcastle, capturing the deformation and compression of the sandcastle under the weight of the book? This scenario presents a significant challenge in achieving realistic image composition—the process of blending a foreground object with a background to create a convincing scene. Although the field has advanced rapidly with developments in diffusion models~\cite{saharia2022photorealistic, dhariwal2021diffusion}, current image composition techniques face several critical limitations. Techniques like Paint-by-Example~\cite{yang2023paint} and ObjectStitch~\cite{song2023objectstitch} allow for image-guided editing of specific scene regions using a target image as a template; however, they fall short in producing identity-consistent content, particularly for categories not covered during training. More importantly, although often appearing to be visually pleasant, these methods neglect the interaction between the inserted object and the background scene—an essential aspect for achieving contextual coherence and true realism. This reveals a deeper shortcoming: a lack of motion- and physics-aware modeling. Without capturing the dynamic and structural response between scenes and inserted objects, current methods fall short of delivering truly realistic image compositions. Overall, diffusion-based image models are purely data-driven and often rely on training distributions, limiting their ability to generate physical realism~\cite{meng2024towards,bastek2024physics,bansal2024videophy,Lin2025ExploringTE}, or handle out-of-distribution scenarios for abstract compositions. In addition, they require manual selection of insertion locations, making them less scalable and more labor-intensive for users. Addressing this gap is essential for advancing beyond static visual alignment toward motion-aware, interaction-consistent synthesis. 

We present InteractPro, a unified framework for motion-aware image composition, integrating three components: an intelligent planner (\textbf{InteractPlan}), a simulation-based composition module (\textbf{InteractPhys}), and a diffusion-based composition module (\textbf{InteractMotion}). InteractPhys models detailed physical interactions using an enhanced Material Point Method (MPM)\cite{jiang2016material}, capturing physical deformation, compression, and structural responses for physically accurate compositions in Fig.~\ref{fig:teaserall}(d-f) or even abstract compositions in Fig.~\ref{atyp} ,which are difficult for current diffusion-based models to accurately replicate. Inspired by PhysGaussian~\cite{xie2024physgaussian} and related physics simulation based methods~\cite{liu2024physics3d,zhang2024physdreamer}, we are the first to extend MPM-based simulation techniques to the field of image composition, marking a key advancement in realism. We adopt Physics3D~\cite{liu2024physics3d} and enhance its control mechanisms for more physically consistent behavior. InteractMotion, built on pretrained video diffusion models, handles complex visual phenomena beyond explicit simulation, generating motion-coherent compositions without extra training, as shown in Fig.~\ref{fig:teaserall}(a-c). Specifically, InteractMotion harnesses motion priors ingrained within Image-to-Video (I2V) diffusion models to endow the inserted foreground objects with dynamic characteristics through a novel motion-aware inpainting, circumventing the necessity for further training and ensures motion coherence with seamless background integration.

Each module compensates for the other's limitations: InteractPhys excels at modeling physically grounded interactions—such as deformation or compression—that are difficult to infer visually, but it is limited in highly complex scenes due to the need for explicit 3D representation and simulation. Conversely, InteractMotion handles visually complex or ambiguous scenes more robustly, but often fails to capture implied physical effects—like compression or weight transfer—due to its lack of deep physical grounding~\cite{Lin2025ExploringTE}. By combining deterministic physical modeling with flexible data-driven synthesis, InteractPro robustly handles both common and out-of-distribution motion-aware composition tasks. To coordinate these strengths, our intelligent planner, InteractPlan, uses GPT-4V~\cite{achiam2023gpt} with multimodal Chain-of-Thought (CoT) prompting~\cite{wei2022chain,zhou2022least,zhang2022automatic}, drawing from LLM-assisted frameworks~\cite{shen2024hugginggpt,li2024sheetcopilot,feng2024layoutgpt,10659157,10480591}. InteractPlan carefully evaluates the characteristics of the foreground and background images, considering multiple factors such as the type of object interaction and environmental dynamics. By balancing a set of thoughtfully crafted criteria, InteractPlan selects the most appropriate method, ensuring optimal results across a diverse range of image scenarios. InteractPlan also automates optimal placement of foregrounds, ensuring logical and visually consistent integration.

Equipped with these techniques, InteractPro excels in achieving motion-aware compositions, as demonstrated in Figure~\ref{fig:teaserall}. It streamlines the planning phase and produces compositions that vividly convey motion and interaction. In summary, our contributions are as follows: 
\begin{itemize}
  \item We pioneer the concept of motion-aware image composition, crafting new scenes from user-defined concepts that inherently capture motion and interaction.
  \item We design \textbf{InteractPlan} with task prompts and CoT reasoning to automate method selection and object placement based on scene dynamics.
  \item We propose \textbf{InteractPhys}, a simulation-based method enabling physically grounded interactions like deformation and compression, with control over surreal behaviors.
  \item We propose \textbf{InteractMotion}, a diffusion-based method that generates dynamically coherence with video motion priors—especially effective in scenarios with complex dynamics.
  \item We evaluate our method against related works on image composition, showcasing its enhanced efficiency in broad scenarios.
\end{itemize}

\begin{figure*}[!t]
\centering
\includegraphics[width=\linewidth]{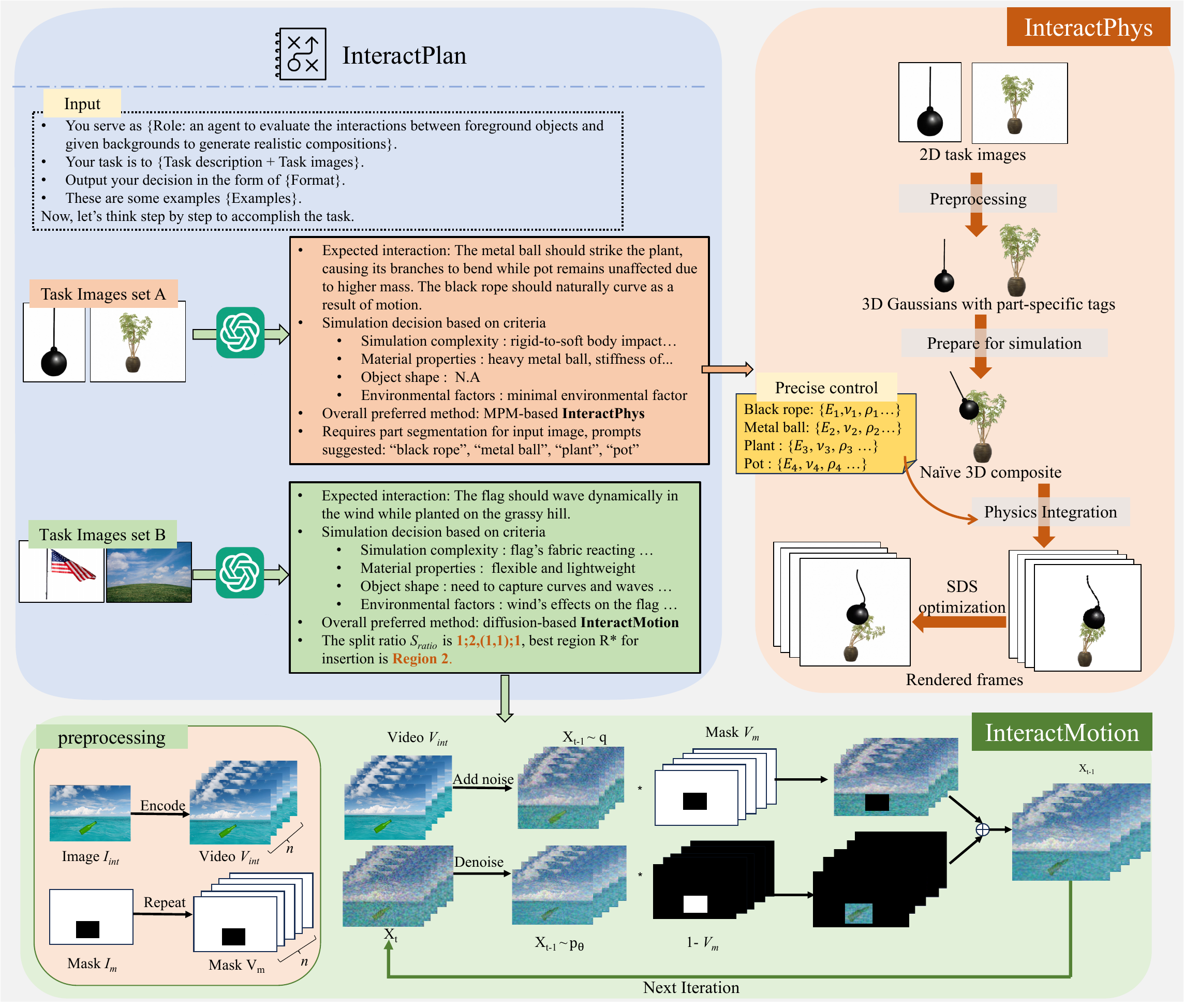}
\caption{The overall pipeline of InteractPro. InteractPlan dynamically determines the most suitable method for object composition, tailored to the specific scenario at hand. InteractPhys offers meticulous control over interactions and object behaviors within the simulation, ensuring that every object interaction strictly adheres to physical laws and delivering physics aware composites. Meanwhile, InteractMotion leverages video priors to produce visually rich effects in scenarios where simulation is insufficient or impractical.}
\label{pipe}
\end{figure*}

\section{Related Works}
\subsection{Generative Image Composition}
Recently, generative image composition techniques~\cite{song2023objectstitch,chen2023anydoor,zhang2023controlcom, yang2023paint,lu2023tf,10480591,hachnochi2023cross,10175586} have sought to address these multifaceted issues with a comprehensive model, enabling end-to-end generation of image-guided composite images. Prominent among these are Paint-by-Example~\cite{yang2023paint}, AnyDoor~\cite{chen2023anydoor}, and ObjectStitch~\cite{song2023objectstitch}, which, despite their progress, struggle with preserving foreground integrity and seamless blending.
Crucially, these methods predominantly result in visually pleasant but static compositions, lacking the dynamic interaction and motion coherence. InteractPro stands out by integrating elements seamlessly—even when they come from vastly different visual domains, such as combining photorealistic backgrounds with cartoon elements. It also introduces context-aware motion and structural responses,allowing inserted objects to interact naturally with their environment and enhancing overall realism.

\subsection{LLM-Assisted Visual Generations}

Large language models (LLMs) like GPT-4 with Vision (GPT-4V)~\cite{achiam2023gpt} excel in various multimodal tasks~\cite{li2022blip,liu2024physics3d,10814093,10017364}. A key development in harnessing the potential of LLMs is the introduction of chain-of-thought (CoT) prompting~\cite{wei2022chain,zhang2023multimodal}, a technique that enhances adaptability to specific tasks. Several approaches has leveraged LLMs as planning component across diverse applications, benefitting from CoT prompting and multimodal reasoning capabilities.~\cite{huang2022language} shows that LLMs behave like zero-shot planners when they are correctly prompted.~\cite{yang2024mastering,10234506} showcase LLMs' planning capabilities by breaking down complex generation tasks into specific subprompts, enabling more precise control over text-to-image and video outputs.~\cite{lv2024gpt4motion,zhou2024scenex} utilise the strong planning capabilities of LLMs for Blender script generation and scene understanding, showcasing their potential in creative and technical domains.

\subsection{Physics Integrated Representations}

In 3D asset creation, physics-based techniques allow the generation of motion driven by physical interactions.~\cite{chen2022virtual} achieves joint reconstruction of geometry, appearance, and physical properties for elastic objects through a multiview capture system using compressed air, enabling realistic animations under new physical conditions.~\cite{li2023pac} and~\cite{xie2024physgaussian} integrate physics simulations with NeRF and 3D Gaussian frameworks, respectively, to produce physically accurate motion. Building on the PhysGaussian~\cite{xie2024physgaussian} approach,~\cite{zhang2024physdreamer} and~\cite{liu2024physics3d} further integrate the optimization of material properties into physics-based 3D Gaussians by leveraging pre-trained video generation models. We extend MPM into image composition task and enhance the capabilities of~\cite{liu2024physics3d} for more physically coherent results.

\section{Method}
\label{sec:formatting}

Existing image composition methods often produce static results where foreground objects appear visually aligned but remain disconnected from their new context. They miss subtle cues that signal interaction, making the composite feel artificial. We propose InteractPro in Fig.~\ref{pipe}, a three-part framework with an intelligent planner (InteractPlan) and two composition modules: a physics-based simulator (InteractPhys) and a diffusion-based composer (InteractMotion), enabling context-aware, dynamic compositions.

\subsection{Preliminary: Physics3D}
InteractPhys adopts the simulation model Physics3D~\cite{liu2024physics3d}, which builds on PhysGaussian~\cite{xie2024physgaussian}—a framework that integrates continuum mechanics with 3D Gaussian Splatting (GS) for generative dynamics. In this setup, physics-integrated 3D Gaussians act as discrete particle clouds, spatially discretizing the continuum. Physics3D extends this by incorporating additional parameters such as viscosity and the Lamé coefficients \( \lambda \) and \( \mu \), improving its ability to model inelastic object behavior. The particle dynamics are governed by a continuum deformation map and simulated via the Material Point Method (MPM), which tracks mass and deformation on a background grid. A brief overview of continuum mechanics and MPM is included in the Supplementary Material. Overall, the parameters are updated with:
\begin{equation}
\nabla_\theta \mathcal{L}_{\text{SDS}} = \mathbb{E}_{t, p, \epsilon} \left[ w(t) \left( \epsilon_\phi(I_t^p; t, I_t^r, \Delta p, y) - \epsilon \right) \frac{\partial I_t^p}{\partial \theta} \right].
\label{eq:sds}
\end{equation}

Here, $w(t)$ represents a time-dependent weighting function, $\epsilon_\phi(\cdot)$ denotes the predicted noise generated by a 2D diffusion prior $\phi$, $\Delta p$ signifies the relative change in camera pose from the reference camera $r$, and $y$ denotes the given condition (i.e., image or text).

\subsection{InteractPlan}
\label{sec:interactplan}

We leverage the reasoning and multimodal capabilities of GPT-4V as our planner, \textbf{InteractPlan}, which intelligently selects between \textbf{InteractPhys} (Section~\ref{sec:InteractPhys}) and \textbf{InteractMotion} (Section~\ref{sec:InteractMotion}) by analyzing object interactions, environmental effects, material properties, and object complexity. This ensures the application of the most suitable method for realistic, seamless compositions across diverse scenarios.

To enable expert-like decision-making, we design a structured prompt template comprising:

\begin{itemize}
    \item \textbf{Role}: The LLM acts as an evaluator of foreground-background interactions. 
    \item \textbf{Task}: The LLM determines the appropriate method by evaluating interaction types (e.g., collisions, compression for InteractPhys; shape deformation, light refraction for InteractMotion), material behaviors (e.g., jelly/sand vs. fluid/surface tension), environmental dynamics (e.g., wind, lighting), and object shape complexity. If InteractPhys is selected, it outputs part-specific segmentation prompts based on material differentiation. If InteractMotion is selected, it predicts a split ratio \( \mathit{S}_{\mathit{ratio}} \) (per~\cite{yang2024mastering}) and identifies the optimal region \( R^* \) in background image \( \mathit{I}_{\mathit{bg}} \) for inserting the foreground object. 
    
    \item \textbf{Format}: A strict output schema ensures consistent communication across the pipeline.
    
    \item \textbf{Examples}: Scenario-specific examples guide the planner in applying criteria to various interaction settings, ensuring correct method selection and format adherence.
\end{itemize}

A full prompt example and Chain-of-Thought (CoT) process are elaborated in the Supplementary. If InteractMotion is chosen, the foreground is then automatically inserted into \( R^* \) of \( \mathit{I}_{\mathit{bg}} \) to generate an intermediate composite \( \mathit{I}_{\mathit{int}} \) and mask \( \mathit{I}_{\mathit{m}} \) for diffusion-based synthesis. 

\subsection{InteractPhys}
\label{sec:InteractPhys}

We show the workflow of InteractPhys in Fig.~\ref{pipe}. In MPM simulation, objects are represented as 3D Gaussians, where each particle’s physical properties—mass \( m \), Young’s modulus \( E \), Poisson’s ratio \( \nu \), Lamé coefficients \( \lambda, \mu \), and viscosity \( v \)—govern their physics-based motion. In particular, \( E \) and \( \nu \) control elastic behavior and deformation under force.

\textbf{Preprocessing.} InteractPhys ensures physics-realistic compositions by simulating object interactions in 3D space, employing the differentiable MLS-MPM simulator~\cite{hu2018moving}. We convert 2D image objects into 3D Gaussians via~\cite{tang2024lgm} for compatibility with the simulation framework. If segmentation is required (as determined by InteractPlan), we apply 3D-aware segmentation~\cite{kerbl3Dgaussians,kirillov2023segment} to partition objects and assign labels for fine-grained control. This step is critical when an object comprises parts made of different materials, requiring separate treatment.

\textbf{Precise physics integration.} Unlike Physics3D~\cite{liu2024physics3d}, which treats the entire scene of Gaussian objects as a single entity, our enhanced InteractPhys enables per-part parameter control. For a scene/object with \( n \) segments, each part \( P_i \) is assigned its own physical parameters \( \{ E_i, \nu_i, \lambda_i, \mu_i, v_i, m_i \} \). These variations allow different materials within an object to exhibit distinct responses to external forces, enhancing realism and physical accuracy.

\textbf{Optimization Step.} Users may optimize physics parameters using Score Distillation Sampling (SDS) loss~\cite{poole2022dreamfusion} and its subsequent works~\cite{wang2023prolificdreamer,yang2025texttoimage,yang2024learn} when manual tuning is insufficient. In~\cite{liu2024physics3d}, material parameters—Young’s modulus \( E \), Lamé coefficients \( \lambda \), \( \mu \), and viscosity \( v \)—were updated independently. However, treating \( \lambda \) and \( \mu \) separately from \( E \) and Poisson’s ratio \( \nu \) breaks physical consistency, often leading to unrealistic stress responses and deformations.

To address this, we reparameterize \( \lambda \) and \( \mu \) in terms of \( E \) and \( \nu \), following~\cite{bonet1997nonlinear}, ensuring physically coherent material behavior:

\begin{equation}
\lambda = \frac{E \nu}{(1 + \nu)(1 - 2\nu)}, \quad
\mu = \frac{E}{2(1 + \nu)}
\label{eq:lame}
\end{equation}

This ensures that volumetric (\( \lambda \)) and shear (\( \mu \)) responses remain grounded in physically valid relationships. Our optimization still uses the gradient formulation in Equation~\ref{eq:sds}, but unlike prior work, derives \( \lambda \) and \( \mu \) analytically to maintain physical realism and ensure material's elastic behavior remains consistent under all deformations. 

\begin{figure}[]
\centering
\includegraphics[width=\linewidth]{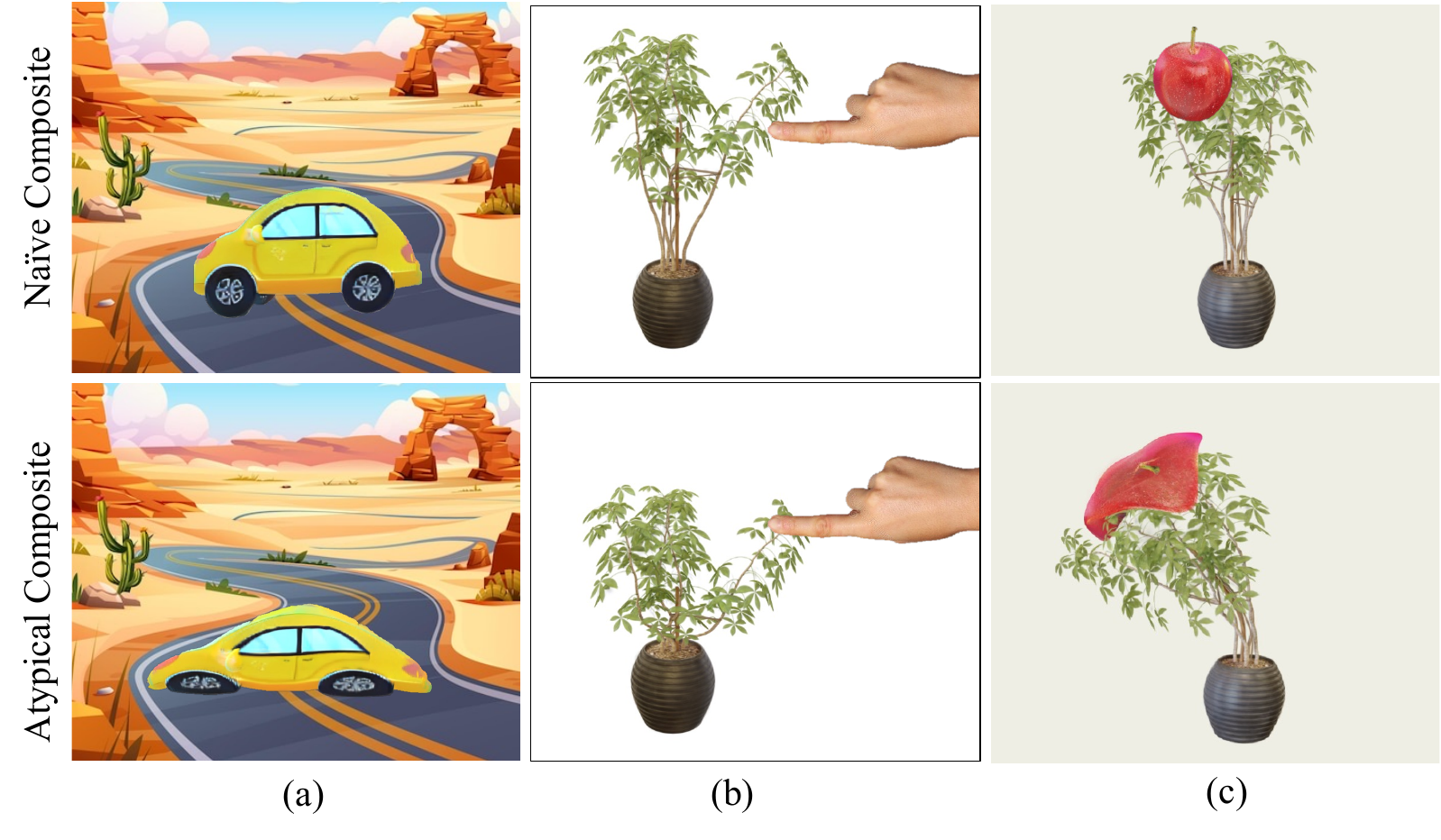}
\caption{Atypical composition with InteractPro. These striking compositions demonstrate the creative potential of InteractPro with surreal visualizations and a testament to the framework’s ability to simulate even the most fantastical scenarios.}
\label{atyp}
\end{figure}

\textbf{Atypical Interactions.} In addition to simulating naturally occurring interactions, InteractPhys excels at generating unique and unconventional compositions. Unlike diffusion-based methods which are constrained by their training data, InteractPhys allows for the application of atypical material properties to familiar objects. For example, an apple can be redefined to behave like slim—falling onto a plant, following its contours, and bending branches under its weight, as shown in Fig.~\ref{atyp}(c). This level of control and flexibility in material manipulation allows for more abstract and physically plausible outcomes, even in scenarios that deviate from standard physics interactions.

\subsection{InteractMotion}
\label{sec:InteractMotion}

\begin{figure*}[]
  \centering
  \includegraphics[width=\textwidth]{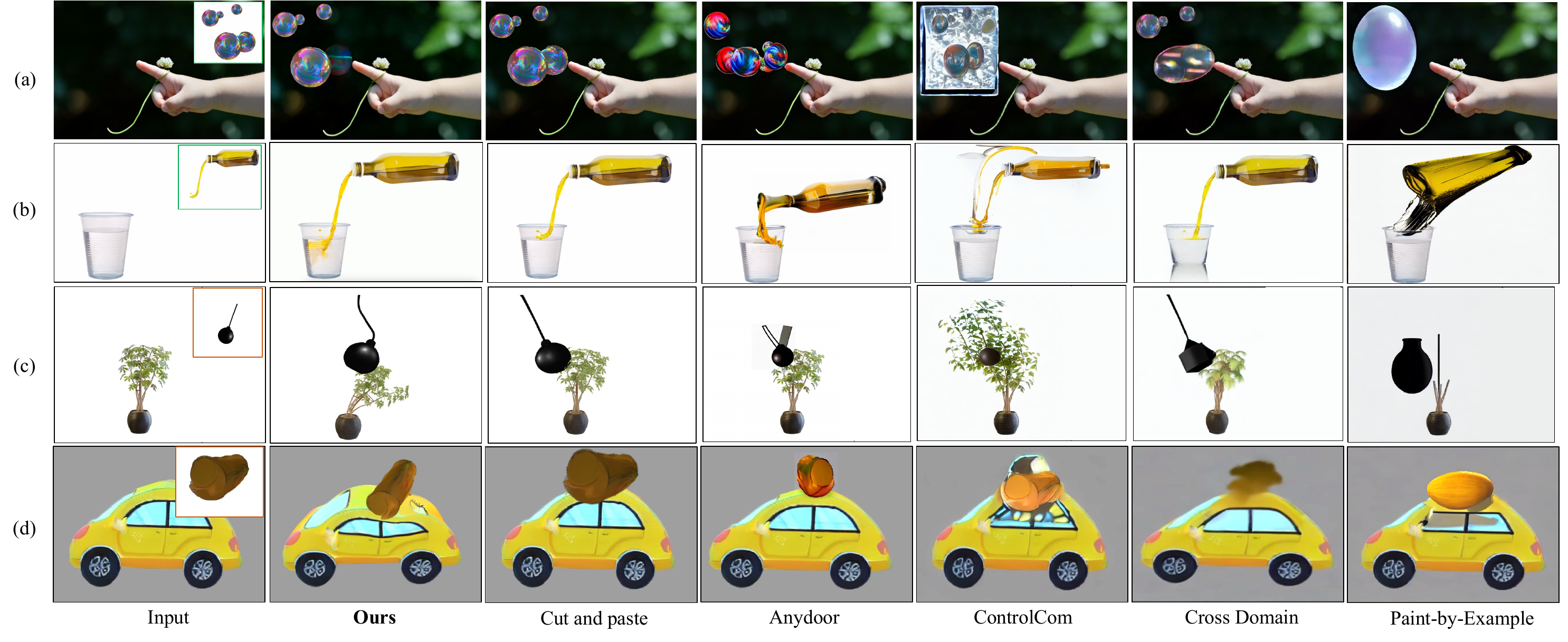}
 
  \caption{Qualitative comparison with existing image composition methods. InteractPro effortlessly harmonize disparate elements into cohesive scenes that align with physical intuition without the need for any additional model training or optimization, while maintaining identity consistency in foreground objects. In contrast, other methods may visually place objects into the scene in an appealing manner, but they neglect the underlying physics and scene context, resulting in compositions fail to hold up under physical scrutiny or real-world dynamics. Rows (a-b) are performed with InteractMotion and (c-d) are with InteractPhys. Please zoom in for better visualizations. See more results in Supplementary.}
  \label{fig:quali_all}
\end{figure*}

\begin{figure*}[]
  \centering
  \includegraphics[width=\linewidth]{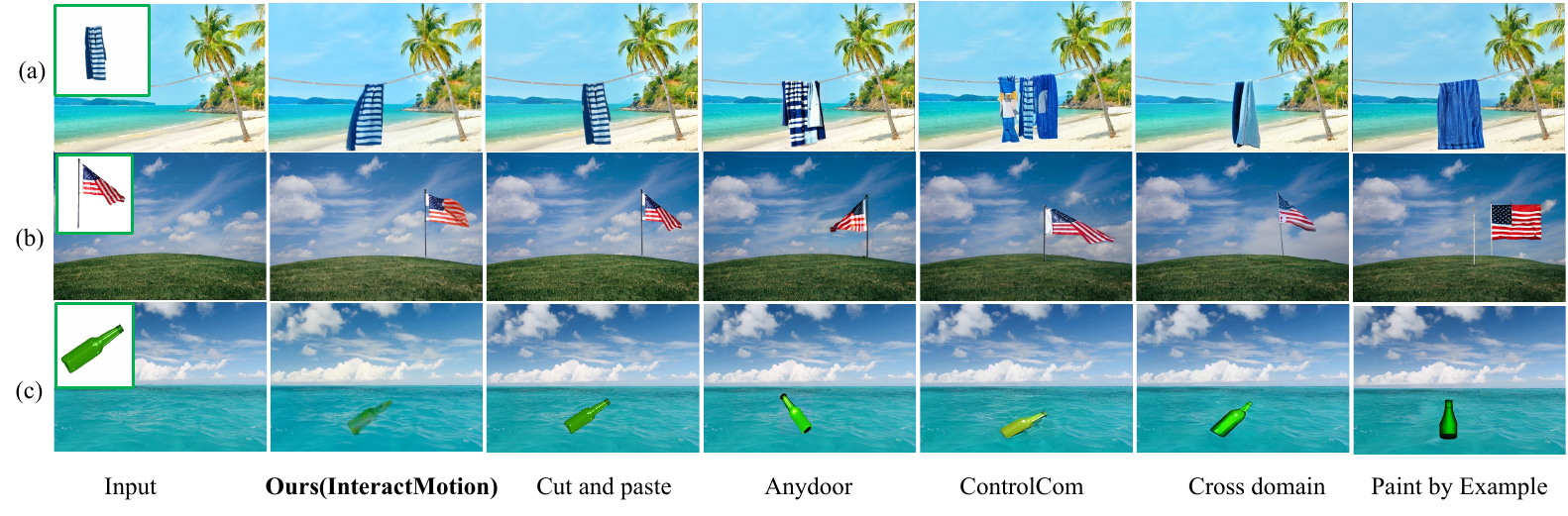}
  \caption{Qualitative comparisons of InteractMotion examples in Fig.~\ref{fig:teaserall}(a-c) with existing methods. Please zoom in for better visualizations.}
  \label{fig:teaser_motionmore}
\end{figure*}

The intermediate composite image \( \mathit{I}_{\mathit{int}} \) from InteractPlan lacks interaction between the inserted object and \( \mathit{I}_{\mathit{bg}} \). To introduce visually coherent, dynamic motion, we propose InteractMotion—which unlocks realistic motion-aware composition by distilling motion priors from pretrained I2V video diffusion models in a zero-shot manner. Given \( \mathit{I}_{\mathit{int}} \) and its mask \( \mathit{I}_{\mathit{m}} \), InteractMotion synthesizes a frame sequence where the inserted object exhibits natural motion while the background remains unchanged. This selective animation is essential for achieving realistic interaction without disrupting the original scene. A naive approach would apply the diffusion model directly to \( \mathit{I}_{\mathit{int}} \), often resulting in unintended background changes or camera motion—contradicting our objective of localized motion. To overcome this, InteractMotion integrates inpainting into the pretrained diffusion process (Fig.~\ref{pipe}), ensuring that only the unmasked object region animates across \( n \) frames, while the masked background remains static.
A physics-aware LLM~\cite{bansal2024videophy} then selects the most appropriate frame as the final motion-aware result.

\textbf{Mask Preprocessing.} The preprocessing of the input mask \( \mathit{I}_{\mathit{m}} \) to generate 
\( \mathit{V}_{\mathit{m}} \) is pivotal for ensuring its consistent application across all \textit{n} frames within a video. This involves duplicating \( \mathit{I}_{\mathit{m}} \) to match the number of frames \textit{n} in the video, creating a temporally extended mask \( \mathit{V}_{\mathit{m}} \) = $\{m^1, m^2, \ldots, m^n\}$
 that is applied identically to each frame, maintaining spatial and temporal consistency throughout the diffusion process of T timesteps. The constancy of \( \mathit{V}_{\mathit{m}} \) across the entire diffusion sequence is a fundamental aspect of our methodology, underpinning the coherence and effectiveness of the generated video content.

\textbf{Background Preserving Inpainting.} We then encode the input \( \mathit{I}_{\mathit{int}} \) to obtain its latent code \( \mathit{V}_{\mathit{int}} \) and \( \mathit{I}_{\mathit{emb}} \) using the VAE encoder~\cite{kingma2013auto} and CLIP Image processor~\cite{radford2021learning} respectively. We start from a randomly initialized latent \( \mathit{X}_{\mathit{t}} \), 
where \( \mathit{X}_{\mathit{t}} = \{x_t^1, x_t^2, \ldots, x_t^n\} \), indicating a video consisting of \( n \) frames at timestep \textit{t}. We use $q$ to denote the diffusion process and $p$ to denote the reverse process. In each timestep, we perform denoising conditioned on \( \mathit{I}_{\mathit{emb}} \) for classifier-free guidance~\cite{ho2022classifier}, yielding a latent denoted \( \mathit{X}_{\mathit{t-1}} \sim p_\theta \). This denoising step specifically addresses the generation of the pixels corresponding to the unknown area—that is, the foreground region designated for inpainting. For each frame \( i \) in \( \mathit{X}_{\mathit{t-1}} \sim p_\theta \) at time step \( t-1 \), it is defined as:
\begin{equation} \label{eq:unknown}
x_{\text{unknown}, t-1}^i \sim \mathcal{N}(\mu_{\theta}(x_t^i, t), \Sigma_{\theta}(x_t^i, t)).
\end{equation}

In addition, we add noise to \( \mathit{V}_{\mathit{int}} \), obtaining its noised version, denoted as \( \mathit{X}_{\mathit{t-1}} \sim q \).  This noising step specifically addresses the reconstruction of the pixels corresponding to the known area—that is, the background region that should be unchanged. For each frame \( i \) in \( \mathit{X}_{\mathit{t-1}} \sim q \) at time step \( t-1 \), the process is defined as:
\begin{equation} \label{eq:known}
x_{\text{known}, t-1}^i \sim \mathcal{N}(\sqrt{\bar{\alpha}_t}x_0^i, (1 - \bar{\alpha}_t)\textbf{I}).
\end{equation}

To maintain background fidelity, we blend the two latents \( \mathit{X}_{\mathit{t-1}} \sim p_\theta \) and \( \mathit{X}_{\mathit{t-1}} \sim q \) using the pre-processed mask \( \mathit{V}_{\mathit{m}} \) to obtain  \( \mathit{X}_{\mathit{t-1}} \). 
The blending is achieved with Equation \ref{eq:combined} to ensure seamless and coherent insertion, which is applied consistently across all \textit{n} frames in the video sequence. The operation uses element-wise multiplication to combine the known and unknown regions, where \( m^i \) refers to the binary mask values applied to the frame $x_{\text{known}, t-1}^i$
to retain the original background, and $x_{\text{unknown}, t-1}^i$ to integrate the denoised foreground. The result, 
 $x_{t-1}^i$is the i\textsuperscript{th} frame of the video \( \mathit{X}_{\mathit{t-1}} \) at timestep \textit{t}-1, which exhibits a coherent and seamless composition of the background and the updated foreground elements:
\begin{equation} \label{eq:combined}
x_{t-1}^i = m^i \odot x_{\text{known}, t-1}^i + (1 - m^i) \odot x_{\text{unknown}, t-1}^i.
\end{equation}

We perform Equation~\ref{eq:combined} for \( T \) sampling steps to generate an inpainted video of \( n \) frames. A physics-aware LLM~\cite{bansal2024videophy} serves as an auto-rater to select the highest-scoring frame as the final motion-aware composite. InteractMotion is also model-agnostic, compatible with any diffusion-based I2V model, allowing seamless integration with evolving video diffusion architectures for continuous improvement.


\section{Experiment}

\subsection{Implementation Details}

We use GPT-4V~\cite{achiam2023gpt} in \textbf{InteractPlan} to handle scene analysis, captioning, and segmentation prompt generation. For segmentation of 3D Gaussian object parts in InteractPhys, we follow~\cite{chen2024gaussianeditor} and apply the 2D Segment Anything model~\cite{kirillov2023segment}. InteractPhys itself is built upon and extends Physics3D~\cite{liu2024physics3d}, allowing finer control over material dynamics. In the optimization process of InteractPhys, we optionally use the ModelScope model `text-to-video-ms-1.7b'~\cite{wang2023modelscope,VideoFusion} to generate 80 frames. For InteractMotion, we use the I2V Stable Video Diffusion (SVD) model~\cite{blattmann2023stable}, producing 25 frames. To enable realistic object movement, we use a loose bounding-box mask rather than a tight one. All images used were sourced from the Internet and combined to form composite inputs for inference. Both modules output video frames where we use a physics-aware LLM~\cite{bansal2024videophy} as an auto-rater to choose the final result.

\subsection{Quantitative Results}

\begin{table*}[h!]
\caption{Quantitative results of methods on DINO and FID. Arrows indicate whether higher or lower values are better. Best values are in bold, and second best values are underlined.}
\centering
\begin{tabular}{l|c|c|c|c|c|c}
\hline
Method & DINO $\uparrow$ & \textbf{DINO-fg} $\uparrow$ & DINO-bg $\uparrow$ & FID $\downarrow$ & FID-fg $\downarrow$ & \textbf{FID-bg} $\downarrow$ \\
\hline
Anydoor~\cite{chen2023anydoor}      & 0.9560 & \underline{0.9457} & \underline{0.9751} & \textbf{20.7703} & \underline{21.8201} & 13.1277 \\
ControlCom~\cite{zhang2023controlcom}   & 0.9318 & 0.9227 & 0.9372 & 32.1804 & 35.2777 & 22.8563 \\
Cross Domain~\cite{hachnochi2023cross} & \underline{0.9613} & 0.9442 & 0.9615 &  22.2624 & 24.8327 & \textbf{9.4748} \\
PbE~\cite{yang2023paint}              & 0.9399 & 0.9346 & 0.9508 & 24.6288 & 30.2789 & 14.3671 \\
Ours                                   & \textbf{0.9726} & \textbf{0.9660} & \textbf{0.9797} & \underline{20.7736} & \textbf{18.4088} & \underline{10.3970} \\
\hline
\end{tabular}
\label{tab:metrics}
\end{table*}

\begin{table}[h!]
\caption{Quantitative results of user study (in percentage).}

\centering
\small
\begin{tabular}{{lcccc}}

\toprule
\multicolumn{1}{p{0.1cm}}{Method} & \multicolumn{1}{p{1.3cm}}{\centering ID \\ Consistency} & \multicolumn{1}{p{0.9cm}}{\centering Seamless \\ Blending } & \multicolumn{1}{p{1.1cm}}{\centering Motion \\ Coherence }  & \multicolumn{1}{p{1.1cm}}{\centering Overall } \\
\midrule 
\midrule


Anydoor~\cite{chen2023anydoor}  & 13.46            & 11.54   & 3.85 & 11.54       \\
ControlCom~\cite{zhang2023controlcom}  & 7.69            & 3.85   & 7.69    & 7.69  \\
CrossDomain~\cite{hachnochi2023cross} & 13.46     & 11.54    & 11.54 &9.62  \\
PbE~\cite{yang2023paint}  & 3.85     & 5.77     & 3.85   &5.77   \\
Ours & \textbf{61.54}     & \textbf{67.31}     &\textbf{73.07 }& \textbf{65.38} 
\\
\bottomrule

\end{tabular}

\label{tab:quan}
\end{table}

For the quantitative evaluations below, we prepare 30 groups of images, each group contains two inputs (foreground and background images) and five outputs generated by the different methods mentioned. Half of the our results are generated with InteractMotion and half with InteractPhys. 

We report DINO~\cite{caron2021emerging} and Fréchet Inception Distance (FID)~\cite{heusel2017gans} scores in Table~\ref{tab:metrics}, distinguishing between foreground and background consistency. The foreground object should remain semantically identifiable while adapting to its environment, so we report DINO-fg as the primary measure of object identity. The background, by contrast, is expected to remain unchanged, making FID-bg the most suitable metric for distributional fidelity. Our method achieves the best performance on DINO-fg, demonstrating strong preservation of object identity, and ranks second on FID-bg, remaining competitive in background fidelity. For completeness and comparability with prior work, we also report DINO-bg, FID-fg, and overall DINO and FID scores. While these auxiliary metrics are less aligned with our task goals, our method remains consistently among the top two across them. 

Importantly, however, while these automatic scores capture background preservation and object identity, they lack the sophistication to differentiate between rudimentary cut-and-paste operations and the more nuanced, motion-aware image compositions characteristic of our work. Recognizing the limitations of traditional evaluation methods, we employ a user survey to measure the impact of our motion-aware compositions. This allows us to gauge user perception, which is pivotal in recognizing the subtleties and dynamism our method infuses into the images, providing a more holistic and suitable evaluation framework for our work. 

We present quantitative comparisons using a user survey on 52 participants with existing related works: Anydoor~\cite{chen2023anydoor}, ControlCom~\cite{zhang2023controlcom}, Cross domain Composition~\cite{hachnochi2023cross} and Paint by Example (PbE)~\cite{yang2023paint}. We provided participants with a standardized Google Forms questionnaire that elaborated on each criterion with definitions and examples to ensure understanding and consistency in responses. Participants are required to select their most preferred method via Multiple Choice Questions, based on object identity consistency, seamless blending, motion aware coherence, and overall harmony considering the above three criteria.  We then collated the votes each method received for each criterion to obtain the quantitative results listed in Table~\ref{tab:quan}, which demonstrates alignment between subjective user ratings and our objective visual analysis.

\subsection{Qualitative Results}


\begin{figure*}[]
  \centering
  \includegraphics[width=\linewidth]{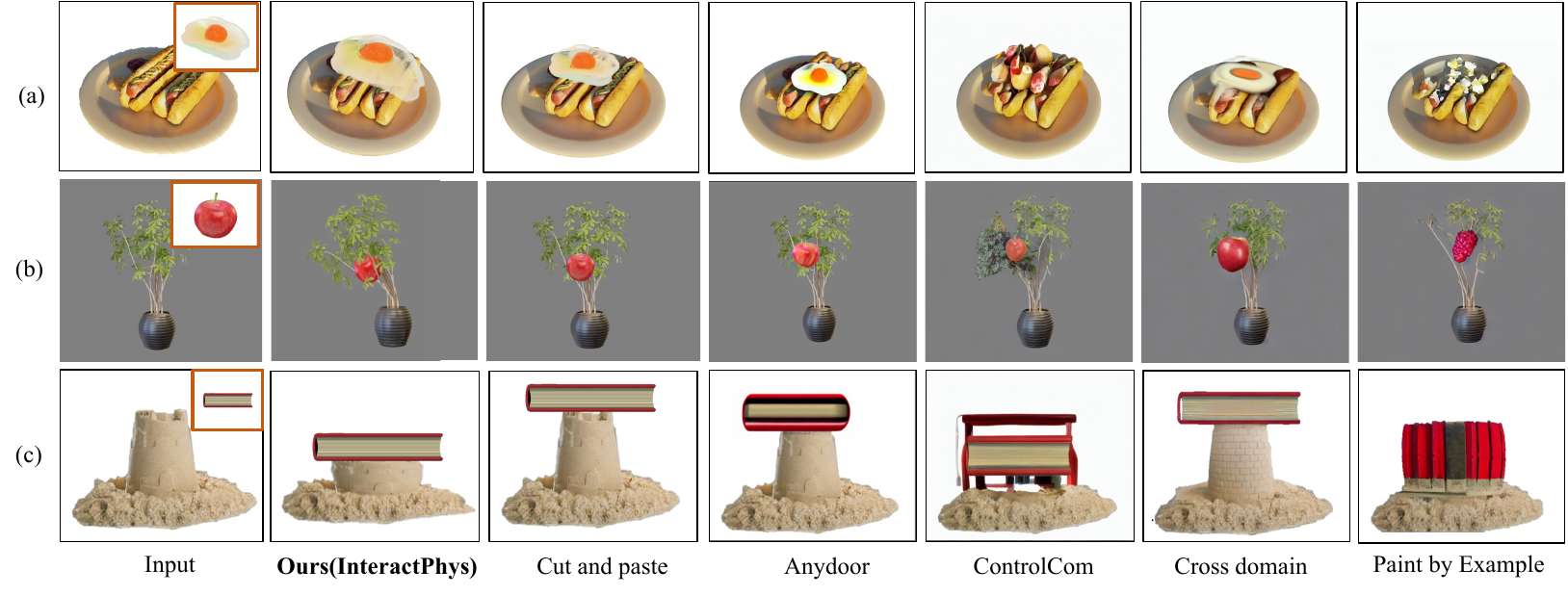}
  \caption{Qualitative comparisons of InteractPhys examples in Fig.~\ref{fig:teaserall}(d-f) with existing methods. Please zoom in for better visualizations.}
  \label{fig:teaser_physmore}
\end{figure*}

\begin{figure*}[!t]
\centering
\includegraphics[width=0.9\linewidth]{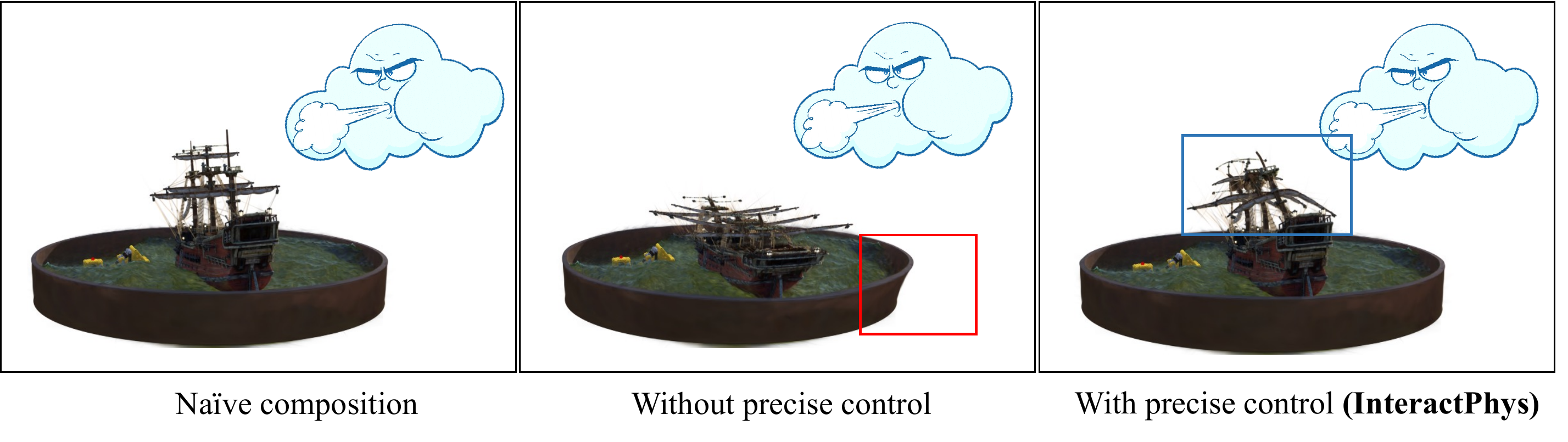}
\caption{Effectiveness of our precise control in InteractPhys. Left: Naive composition. Center: No part segmentation, leading to uniform physics parameters and unrealistic response. Right: InteractPhys with part segmentation, enabling realistic interactions where the ship’s flag is bending with the wind.}
\label{control}
\end{figure*}

\begin{figure}[]
\centering
\includegraphics[width=\linewidth]{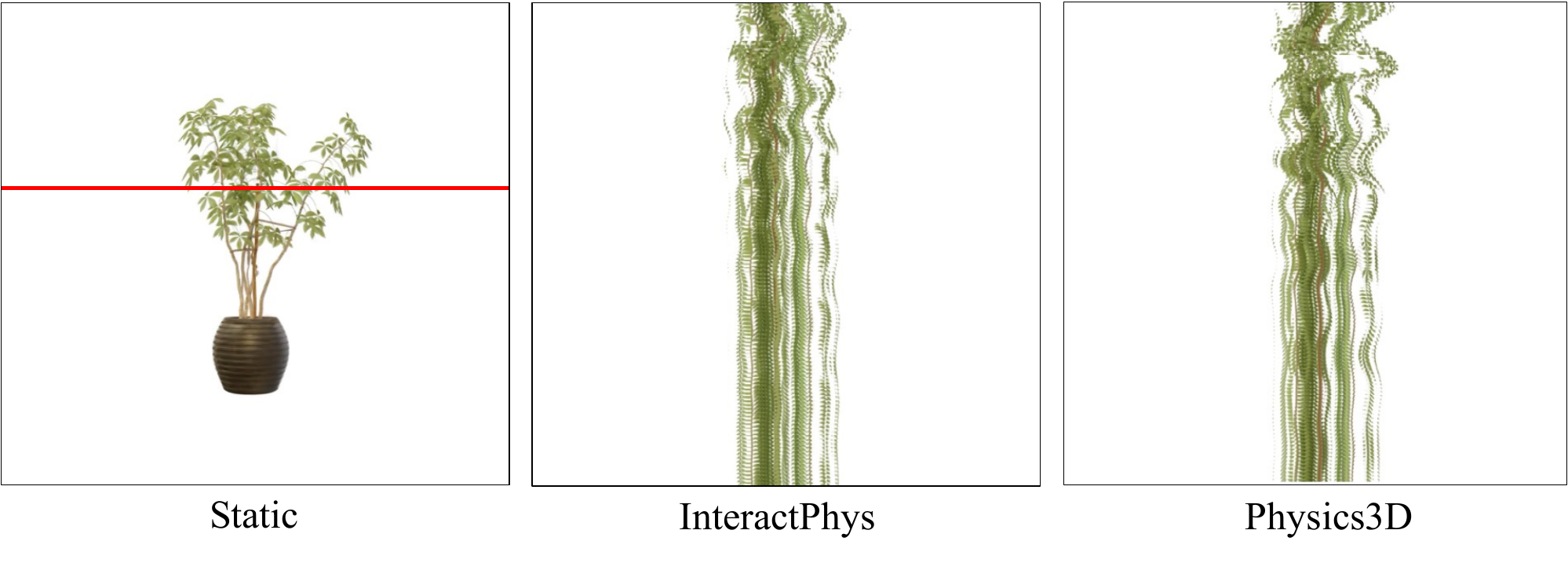}
\caption{Effect of enhanced optimization. Our InteractPhys creates smoother and well controlled motion as compared to Physics3D.}
\label{sds}
\vspace{-1mm} 
\end{figure}


We present qualitative results with the baseline works in Fig.~\ref{fig:quali_all} and a detailed explanation on our composite results below. In (a), introducing a few bubbles near a thumb demonstrates dynamic interaction, where the bubbles burst upon contact with the thumb, capturing a moment of delicate rupture. In (b), a cup of water is mixed with a yellow-colored liquid. The resulting composite vividly illustrates the water adopting traces of yellow to indicate the mixture. In (c), only the plant’s branches bend upon impact of the falling object, demonstrating precise, localized force control—while the pot remains unchanged due to its higher mass. This level of targeted physical deformation is difficult to achieve with diffusion-based methods, which often fail to convey the impact of heavy objects on specific parts of a scene. In (d), the car deforms under the weight of the heavy log, illustrating realistic bending behavior consistent with the material properties and physical context. 

Notably, our method excels in generating motion-aware composites with seamless integration of dynamic foreground objects across various scenes. Analyzing identity consistency, the foreground elements retain their defining characteristics after our InteractMotion composition. In contrast, other methods fail to retain the characteristics of the reference foreground image. For example in row (a) and (d), all four other methods generates bubbles and log respectively, that are different from the reference image. Furthermore, our blending is executed skillfully, leaving no discernible edges or mismatched textures. Most importantly, other methods often overlook the finer physics details of object interactions and lack physical realism. They fail to account for context-driven dynamics like impact of weight in row (c-d), which are crucial for accuracy. Our results underscore the significance of context-aware and physics-driven composition, where inserted objects not only fit visually but also behave according to real-world physical principles, a level of realism that prior methods struggle to consistently achieve.

Fig.~\ref{atyp} highlights the novel simulation capabilities of InteractPro with engaging and non-traditional scenarios. In panel (a), a visualization captures a car melting under the harsh desert sun, depicting the effects of exaggerated heat. Panel (b) depicts a simulated plant that appears ordinary but exhibits an unusual response: it shrinks upon contact, mimicking the behavior of sensitive plants. In (c), apple transforms into a pile of slim like substance, settling onto the plant and causing it to lean due to the redistributed weight. The base of the pot remains unaffected, anchored by its higher mass. These examples highlight the framework's potential to apply physical principles creatively, thereby expanding the boundaries of physical interactions in image compositions that cannot be generalized by diffusion-based models. Note that we are unable to provide qualitative comparison of baselines against the atypical interaction shown in Fig.~\ref{atyp}, as some baselines cannot incorporate additional conditioning—such as text prompts—and can only produce default or typical outputs.

\subsection{More explanation of Fig.~\ref{fig:teaserall} InteractMotion.} We provide explanation of our motion aware composition InteractMotion in Fig.~\ref{fig:teaserall}(a-c), with qualitative comparison in Fig.~\ref{fig:teaser_motionmore}. Our method showcases realistic interactions between inserted foreground object and object in given background. In (a), a towel is placed on a rope at the beach, where the image captures the towel swaying in the breeze. Our method inherently incorporates scale and rotation transformations, guiding the towel’s natural motion in response to the wind. Similarly in (b), the flag is inserted into the open air grass field, where wind presence is likely, shows the flag swaying naturally in the air. In (c), placing a glass bottle into the ocean gives seamless integration. InteractMotion effectively positions the bottle within the sea layer, avoiding the disjointed appearance typical of Cut and Paste, or inconsistent identity in other baselines. The realistic distortion from light refraction in water further enhances this effect. 

\subsection{More explanation of Fig.~\ref{fig:teaserall} InteractPhys.} We provide explanation of our physics aware composition InteractPhys in Fig.~\ref{fig:teaserall}(d-f). In 1(d), the runny egg smoothly spreads over the contours of the hotdog following the shape of hotdog. In 1(e), the branch and leaves of the plant bend under the impact of an apple falling \textit{into} the plant, while the heavier pot remains stationary. In 1(f), the sandcastle is compressed due to the weight of the book. We also show comparisons with other baselines for these cases in Fig.~\ref{fig:teaser_physmore}. Other methods are unable to demonstrate the smooth flow of runny egg against the hotdog, or changes the appearance of input egg. For the apple example, others only show fruit hanging on the plant and seems like on different image layer or looking like cut and paste, whereas ours perfectly place the apple in between the branches of the plant, causing the branches to tilt. Other methods are also unable to demonstrate the weight impact of the book, which should cause the sandcastle to be compressed.

\subsection{Ablation}

\begin{figure*}[]
  \centering
  \includegraphics[width=\textwidth]{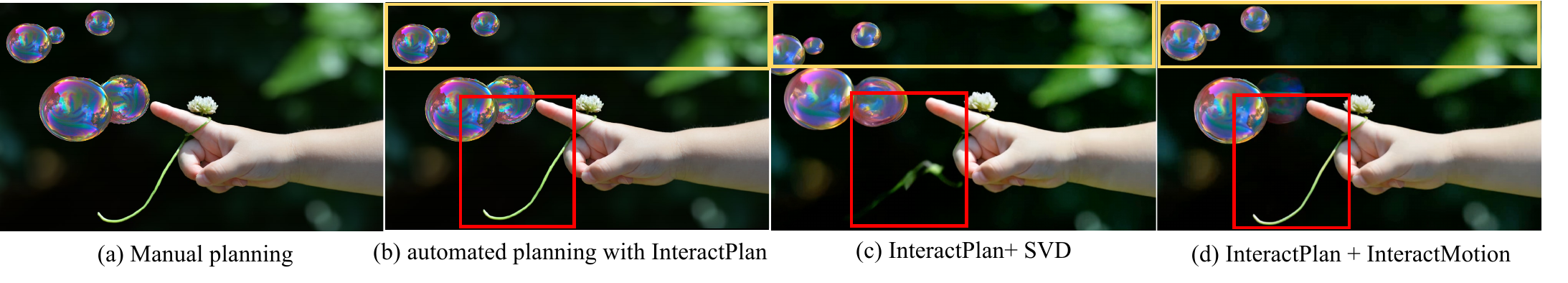}
  \caption{Ablation of our core components. Figures (a) and (b) demonstrate minimal visual differences, yet component (b) could significantly streamlines the workflow. Figure (d) utilizes our proposed InteractMotion that effectively preserves the background scene and camera viewpoint, in contrast to Figure (c) where the background scene (red box) and camera viewpoint (yellow box) exhibits changes.}
  \label{fig:ablation}
  \vspace{-1mm} 
\end{figure*}
 
\textbf{InteractPhys: Effectiveness of precise control.} We illustrate the benefits of part segmentation and per-part control in Fig.~\ref{control}. The left image shows a naive composition where the ship remains static, unresponsive to environmental forces. The center image, without part segmentation, applies uniform physics parameters across all components (ship, water, container), leading to unrealistic effects—such as the container (red box) deforming despite its expected rigidity. In contrast, the right image demonstrates InteractPhys with part-aware control. Segmented components are assigned distinct physics parameters, allowing realistic behaviors: the ship’s flag bends naturally in the wind (blue box), while the rest remains unaffected. This highlights how InteractPhys enables nuanced, physically plausible interactions for complex scenes.

\textbf{InteractPhys: Improved Optimization.} Fig.~\ref{sds} shows space-time slices of simulated video frames to illustrate motion dynamics over time (vertical axis) and space (horizontal axis, red line in “Static” view). The baseline~\cite{liu2024physics3d} suffers from unnatural oscillations due to independently updated Lamé parameters, causing exaggerated plant motion inconsistent with real-world behavior. Our modified approach constrains these parameters by deriving them from physical laws, ensuring more realistic stress responses. As shown in the middle panel (“InteractPhys”), the plant exhibits smoother, more plausible motion, demonstrating improved stability and physical accuracy in the simulation.

\textbf{InteractPlan and InteractMotion.} Fig.~\ref{fig:ablation} evaluate our InteractPlan (applies to both InteractPhys and InteractMotion), and InteractMotion. For object placement planning, we compare manual cut-and-paste (a) with our InteractPlan (b), which automates object placement comparable to human intuition. To assess InteractMotion, we compare generation results using direct Stable Video Diffusion (SVD) in (c) versus our inpainting-based method in (d). InteractMotion better preserves background consistency while introducing localized, realistic foreground motion.

\section{Conclusion}

We present InteractPro, a unified framework for motion-aware image composition that bridges physically grounded simulation and data-driven motion synthesis. By combining a planner-guided system with two complementary modules- InteractPhys for explicit physics simulation and InteractMotion for leveraging motion priors from video diffusion models-InteractPro enables realistic and controllable object-scene interactions across broad range of scenarios. Our results show that this hybrid approach effectively addresses limitations in existing methods, producing dynamic compositions that are visually coherent and physically plausible. 

\textbf{Limitations and Future Works.} While InteractPro provides a modular framework for physics-aware visual composition, each component has limitations. InteractPlan currently uses heuristic decision rules and operates in a single-agent setting, which can lead to suboptimal choices in ambiguous or multi-object scenarios. Future work could incorporate learning-based or multi-agent planners with visual reasoning to improve robustness and scalability. InteractPhys, though enabling controllable physical interactions, is restricted to scenes and objects that can be reconstructed in 3D and presently supports only a limited set of simulation materials. Extending it to partial geometry, learning-based material inference, or hybrid 2D–3D simulation would broaden its applicability. InteractMotion inherits variability from its underlying video diffusion prior, leading to differences across seeds. Future directions include flow-based guidance, motion regularization, or latent alignment constraints to improve consistency.




 \bibliographystyle{IEEEtran}

 \bibliography{main}
%

\newpage

\include{supp}

\vfill

\end{document}

%% file: supp.tex
\section*{Supplementary Material for InteractPro}

\section{More Preliminary}

\textbf{Continuum mechanics} describes the motion of materials through a deformation map \( \mathbf{x} = \phi(\mathbf{X}, t) \), which maps the material space \( \Omega^0 \) to the world space \( \Omega^n \) \cite{bonet1997nonlinear}. The deformation gradient \( \mathbf{F} = \frac{\partial \mathbf{x}}{\partial \mathbf{X}} \) captures local rotations and strains. For viscoelastic materials, the elastoplastic and viscoelastic components \( \mathbf{F}_E \mathbf{F}_P \) and \( \mathbf{F}_N \mathbf{F}_V \) combine in parallel as:
\begin{equation}
\mathbf{F} = \mathbf{F}_E \mathbf{F}_P = \mathbf{F}_N \mathbf{F}_V.
\label{eq:deformation_gradient}
\end{equation}
In the Physics3D \cite{liu2024physics3d} framework, materials are modeled with two parallel components, but only the elastic parts \( \mathbf{F}_E \) and \( \mathbf{F}_N \) contribute to the internal stresses \( \sigma_E \) and \( \sigma_N \).

The system is expressed with dynamic equations. For velocity field \( \mathbf{v(x,t)} \) and density field \( \rho(\mathbf{x}, t) \), the conservation of momentum and mass \cite{germain1998functional} are given by:
\begin{equation}
\rho \frac{D\mathbf{v}}{Dt} = \nabla \cdot \sigma + \mathbf{f}, \quad \frac{D \rho}{Dt} + \rho \nabla \cdot \mathbf{v} = 0.
\label{eq:conservation_laws}
\end{equation}
Here, \( \mathbf{f} \) is the external force, and \( \sigma = \sigma_E + \sigma_N \) is the total internal stress. The strain tensor is updated after computing the material point.

\textbf{Material Point Method (MPM)} discretizes materials into particles, allowing the complete history of strain and stress to be tracked using a particle-to-grid (P2G) and grid-to-particle (G2P) transfer process. This technique has proven effective for simulating various viscoelastic and viscoplastic materials \cite{yue2015continuum, ram2015material}. In MPM, mass and momentum are transferred from particles to grids during P2G as follows:
\begin{equation}
m_i^n = \sum_p w_{ip}^n m_p, \quad m_i^n \mathbf{v}_i^n = \sum_p w_{ip}^n m_p (\mathbf{v}_p^n + \mathbf{C}_p^n (\mathbf{x}_i - \mathbf{x}_p^n)),
\label{eq:mpm_mass_momentum}
\end{equation}
where \( i \) and \( p \) represent grid points and particles, respectively. Each particle \( p \) carries properties such as volume \( V_p \), mass \( m_p \), position \( \mathbf{x}_p^n \), and velocity \( \mathbf{v}_p^n \). After P2G, the updated velocity on the grid is:
\begin{equation}
\mathbf{v}_i^{n+1} = \mathbf{v}_i^n - \frac{\Delta t}{m_i} \sum_p \tau_p^n \nabla w_{ip}^n V_p^0 + \Delta t \mathbf{g},
\label{eq:mpm_velocity}
\end{equation}
where \( \mathbf{g} \) is the gravitational acceleration. G2P then transfers the velocities back to particles and updates the particle stress using the Kirchhoff stress tensor:
\begin{equation}
\tau_p^{n+1} = \tau (\mathbf{F}_E^{n+1}, \mathbf{F}_N^{n+1}),
\label{eq:mpm_stress}
\end{equation}
where \( \mathbf{F}_E^{n+1} \) and \( \mathbf{F}_N^{n+1} \) are components of the strain tensor. This method enhances MPM’s ability to generalize across a wide range of material types, including those found in real-world simulations.

\section{InteractPlan Details}

\subsection{Full prompt template for method decision}

You serve as \textbf{Role:} an agent to evaluate the interactions between foreground objects and given background image to generate realistic compositions. Your primary goal is to analyze and determine which simulation method best suits the interaction scenario, based on the simulation complexity, object’s material properties, environmental effects, and object shape.

Your task is to \textbf{Task description:} evaluate the possible interaction between the foreground object(s) and background. Based on the physical interaction types, material properties, environmental factors, and object shapes, you must select the appropriate method for simulation, i.e., InteractPhys for collision, compression, and deformation, or InteractMotion for complex shape changes and light refraction. 

Output your decision in the form of \textbf{Format:} structured in the following manner:

\begin{itemize}
    \item Expected interaction: A description of how the objects could possibly interact.
    \item Simulation decision based on criteria: Detailed reasoning that help you to decide which method works best.
    \begin{itemize}
        \item Simulation complexity: Level of complexity involved in simulating the interaction.
        \item Material properties: Overview of material properties (e.g., elasticity, fluid dynamics, surface tension).
        \item Object shape: Consideration of the objects' geometries, whether simple or complex.
        \item Environmental factors: Factors like wind, light, gravity, or other external forces.
    \end{itemize}
    
    \item Overall preferred method: The method selected (InteractPhys or InteractMotion).
    \item If InteractPhys chosen: Evaluate if part segmentation is required for input image. If yes, suggest prompts for the segmentation: A list of prompts for segmenting relevant parts of the input image, such as “object 1”, “object 2”. 
    \item If InteractMotion chosen: Evaluate the optimal split ratio of the background image and optimal insertion region of foreground object: The split ratio Sratio is (x,y); best region R* for insertion is Region Z. The evaluation is based on this process: Section \ref{subsec:objplace}.
\end{itemize}

\noindent\rule{\linewidth}{0.4pt}

\begin{figure}[]
  \centering
  \includegraphics[width=\linewidth]{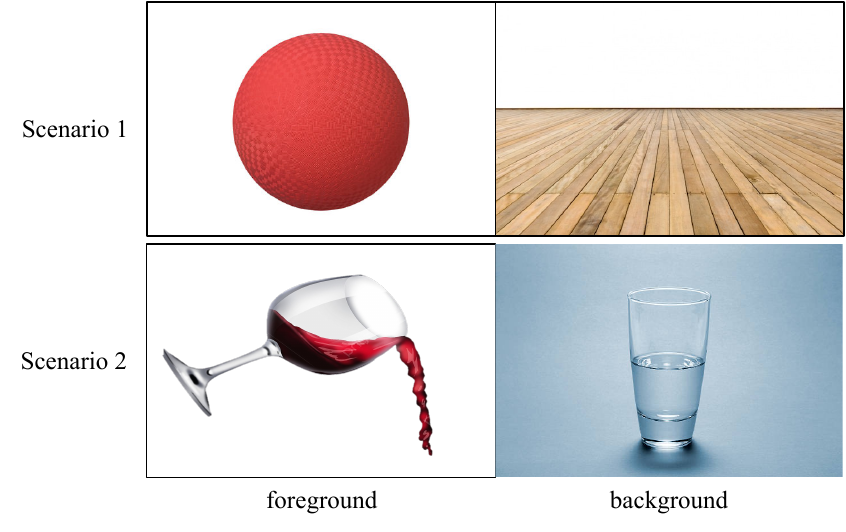}
  \caption{Image sets for examples to InteractPlan.}
  \label{fig:plan_supp}
 
\end{figure}

These are some examples and images are in Fig.\ref{fig:plan_supp}:

\textbf{Scenario 1:} A rubber ball (foreground), and a wooden floor (background).

\textbf{Expected interaction:} The ball will compress upon impact with the wooden surface, deform slightly due to its elastic properties.

\begin{itemize}
    \item Simulation decision based on criteria below:
    \begin{itemize}
        \item Simulation complexity: Elastic deformation and collision dynamics. Moderate.
        \item Material properties: Rubber ball is soft, highly elastic; undergoes noticeable deformation on impact. Wood floor is rigid, assumed non-deformable for simplification.
        \item Object shape: Ball - geometrically simple and symmetrical; simplifies collision detection and deformation modeling.
        \item Environmental factors: N.A.
    \end{itemize}
    
    \item \textbf{Overall preferred method: MPM-based InteractPhys.}
    \item Requires part segmentation for input image, prompts suggested: "rubber ball", "wood surface".
\end{itemize}

\noindent\rule{\linewidth}{0.4pt}

\textbf{Scenario 2:} Wine pouring from glass wine (foreground), and a static glass of water (background). 

\textbf{Expected interaction:} The wine creates ripples across the water surface and forms swirling patterns of mixed colors.
\begin{itemize}
    \item Simulation decision based on criteria below:
    \begin{itemize}
        \item Simulation complexity: Involves fluid transfer between two containers, surface ripple generation, color diffusion, and complex fluid-fluid interaction. Hard.
        \item Material properties: Wine and water — slight density and viscosity difference, both exhibit surface tension. Total 2 liquids and 2 containers.
        \item Object shape: Wine glass and water glass — two distinct container geometries. 
        \item Environmental factors: Gravity affects water flow; surface tension forces present.
    \end{itemize}
    \item Overall preferred method: Particle-based InteractMotion.
    \item The split ratio Sratio is 1,(1,1); 2, best region R* for insertion is Region 0.
\end{itemize}

\noindent\rule{\linewidth}{0.4pt}

Below are a general set of rules.
\begin{itemize}
  \item \textbf{Simulation Complexity:}
    InteractPhys is for localized physical interactions such as collisions, elastic/plastic deformation, and rigid or semi-rigid body contact. Typically used when it is easy to perform simulation. InteractMotion is preferred for large-scale transformations, complex topology changes, continuous flows, or phenomena involving optical or dynamic surface behavior. Typically used when it is hard to perform simulation.

  \item \textbf{Material properties:}
     InteractPhys excels with granular or deformable materials such as sand or jelly.
      InteractMotion handles flow-like effects and phenomena such as surface tension. Typically in fluids and gases.

  \item \textbf{Environmental effects:} InteractMotion is preferred if mechnical forces such as impact or pressure dominate. InteractPhys is preferred if factors like wind, light dynamics are present.

  \item \textbf{Object shape and structure:} Simple and uniform shapes can be handled by both modules. Complex, dynamic shapes favor InteractMotion.

\end{itemize}

Now, let’s think step by step to accomplish the task.
\noindent
\makebox[\linewidth]{\leaders\hrule height 0.4pt\hfill \ END OF PROMPT\ \ \leaders\hrule height 0.4pt\hfill}

\subsection{Object Placement Decision For InteractMotion}
\label{subsec:objplace}

\begin{figure*}[]
  \centering
  \includegraphics[width=\linewidth]{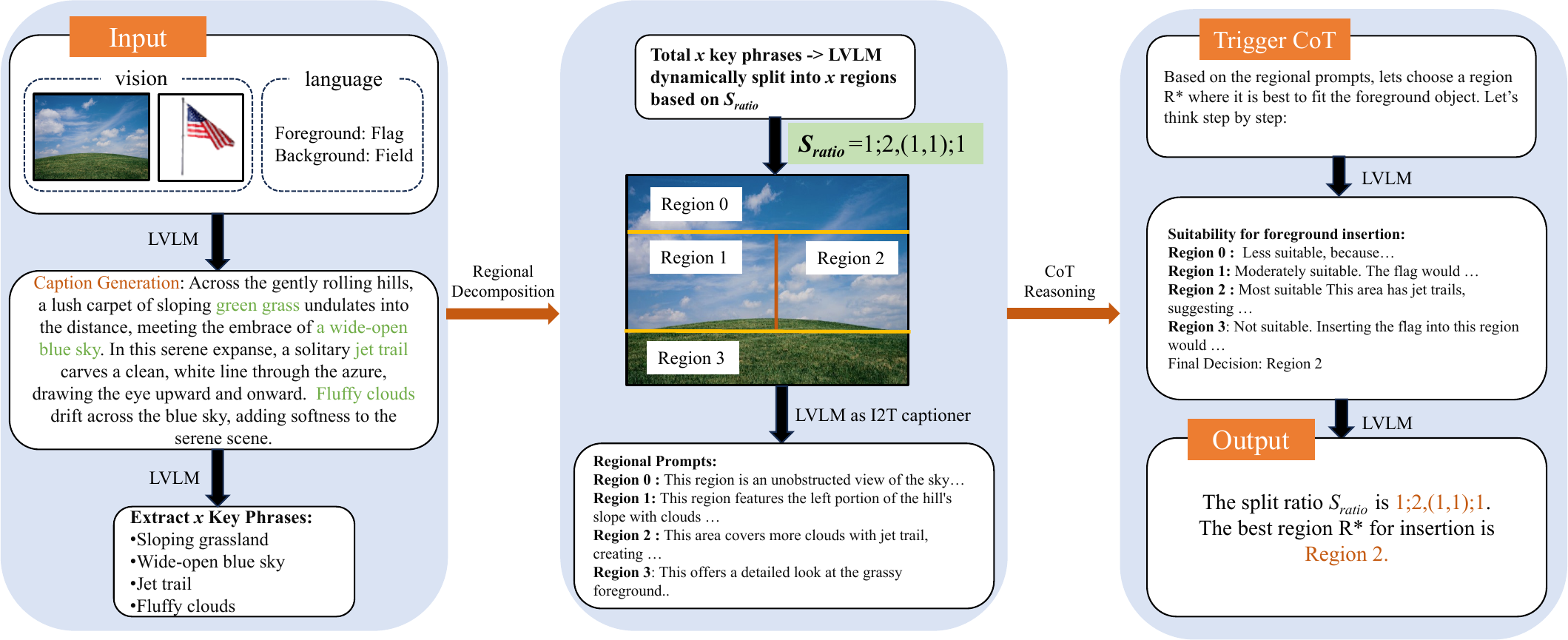}
  \caption{Overview of InteractPlan automated planning phase. When given a set of images with concise prompts, we utilise LVLM to output the split ratio and the most suited region for foreground insertion. Given these outputs, we can automatically create the intermediate composites. }
  \label{fig:gpt}
\end{figure*}

To automate the object placement for InteratMotion, InteractPlan employ a multi-modal LVLM, specifically GPT-4V, to strategically determine the insertion region. This planning phase is partly inspired by \cite{yang2024mastering} where such planning is for image generation, and we further adapt to yield an initial composite where the foreground object is ideally situated to suggest possible motion. We show the thought process of InteractPlan in Fig.~\ref{fig:gpt}.

\textbf{Image Captioning and Regional Decomposition.} We initiate our process by generating a comprehensive caption that integrates the initial concise caption with the visual content of the \( \mathit{I}_{\mathit{bg}} \), leveraging the multimodal capabilities of LVLM to blend textual and visual information. Following this, the LVLM extracts \textit{x} pivotal key phrases from the enriched caption, which in turn inform the dynamic segmentation of the background into multiple regions. This segmentation process is governed by a split ratio, \(\mathit{S}_{\mathit{ratio}} \), dynamically determined by the LVLM to optimally accommodate the identified key phrases. Leveraging the LVLM as image-to-text (I2T) captioner, each region is then assigned a subprompt, meticulously crafted by the LVLM to provide detailed and informative descriptions specific to that segment. This layered approach, from initial comprehensive captioning to detailed regional subprompts, lays a solid foundation for the nuanced planning required in the subsequent phases of image composition.

\textbf{Chain-of-Thought Reasoning.} Building upon the detailed regional prompts derived from the segmentation of \( \mathit{I}_{\mathit{bg}} \), we employ the Chain-of-Thought (CoT) reasoning capabilities of the LVLM \cite{zhang2023multimodal} to meticulously evaluate each region \(R_i\) where \( R_i \subseteq \{R_1, R_2, \ldots, R_x\} \). This evaluation is informed by the specific descriptions provided for each segment, enabling the LVLM to judiciously select the most suitable region for the insertion of the foreground object. The depth and specificity of these regional prompts thus play a critical role, furnishing the LVLM with the contextual insights needed to make an informed decision that optimally aligns the foreground object within the dynamic tapestry of the background scene.
To guide the selection of the optimal region \textit{R*} for the insertion of the foreground object, we adhere to three pivotal criteria while crafting in-context examples and generating detailed rationales. First, there must be sufficient room in the chosen region to accommodate the foreground object, ensuring minimal obstruction of the background scene’s key features. Second, the placement of the foreground object should be in an area that enables plausible motion-based interactions with specific elements in the background conducive to motion, rather than merely the largest objects present. Lastly, we give preference to background regions rich in elements that naturally facilitate or enhance the perceived motion of the foreground object, thereby enriching the dynamic interaction within the composition.

 \section{More Implementation Details}

The InteractPlan module completes in 10 seconds. The image generated by InteractMotion using SVD is rendered at a resolution of 576 $\times$ 1024, whereas the output from the text-to-video-ms-1.7b model (InteractPhys) is generated at 1080 $\times$ 1920 resolution. During InteractMotion, all intermediate composite images from the planning phase and their corresponding masks are reshaped to 576 $\times$ 1024. Instead of using precise object masks, a general bounding box is employed to allow flexibility for object movement in motion-aware composition. InteractMotion takes approximately 90 seconds and consumes 10 GB of memory on a single Tesla V100 GPU with $T = 25$ sampling steps. InteractPhys, on the other hand, takes 60 seconds and uses 15 GB of memory on a single Tesla V100 to render 80 frames, without parameter optimization. With optimization enabled, memory usage increases to 17 GB, and training takes about 40 seconds per epoch, followed by 60 seconds for final rendering. Typically, the system is optimized with 10 epochs.

\begin{figure*}[]
  \centering
  \includegraphics[width=\linewidth]{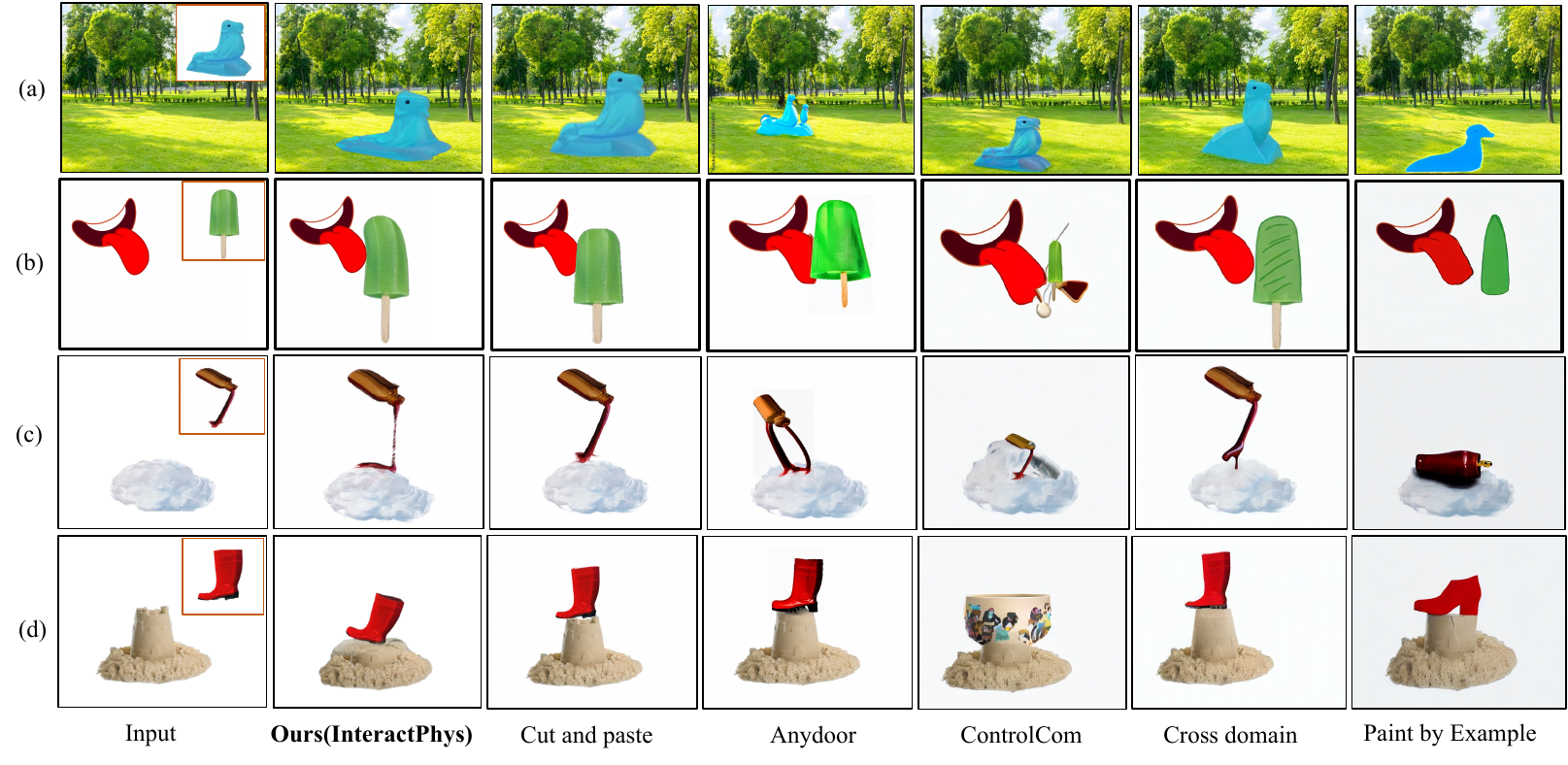}
  \caption{More qualitative comparisons of InteractPhys with existing methods. Please zoom in for better visualizations.}
  \label{fig:phys_more}
\end{figure*}

\begin{figure*}[]
  \centering
  \includegraphics[width=\linewidth]{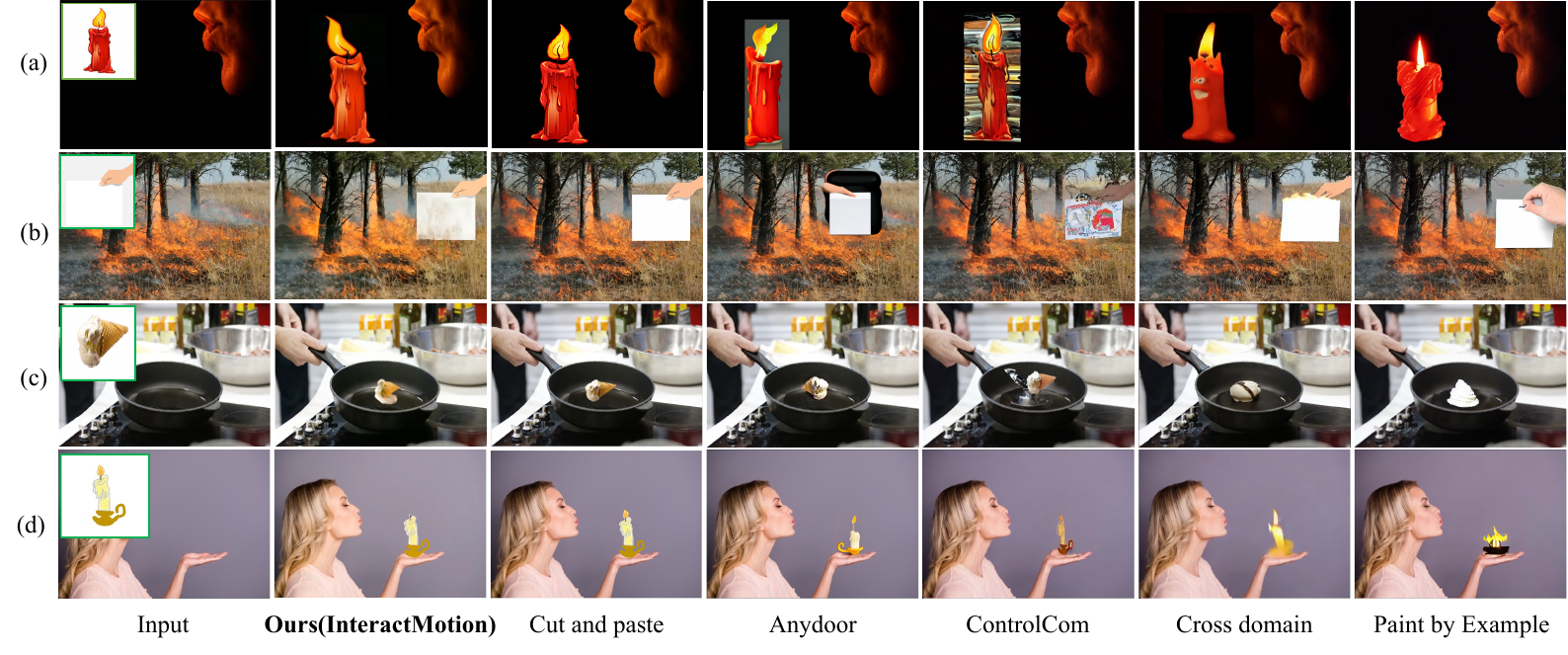}
  \caption{More qualitative comparisons of InteractMotion with existing methods. Please zoom in for better visualizations.}
  \label{fig:motion_more}
\end{figure*}

\section{Extra Qualitative Results for InteractPhys}

We provide additional qualitative comparisons in Fig. \ref{fig:phys_more}, showcasing our InteractPhys against prior methods. While other approaches, such as ControlCom and Paint-by-Example (PbE) in rows (b) and (d), occasionally produce composites with identity inconsistencies, our method consistently maintains object identity without alteration. Although existing methods can create visually appealing compositions, they often overlook the finer physics details of object interactions and lack physical realism. For instance, they fail to account for context-driven dynamics like thermal changes and impact of weight, which are crucial for accuracy. In contrast, our method captures these effects, as seen with the melting ice sculpture in sunny park in (a), and tilting reaction of ice cream to force of tongue in (b). In (c), the liquid flow follows the contours of the fluffy cloud-shaped cotton, demonstrating natural fluid dynamics. In (d), the sandcastle deforms and collapse under the weight of the red boot, illustrating realistic behavior consistent with the material properties and physical context. These results underscore the significance of context-aware and physics-driven composition, where inserted objects not only fit visually but also behave according to real-world physical principles, a level of realism that prior methods struggle to consistently achieve.

\section{Extra Qualitative Results for InteractMotion}

We provide additional qualitative comparisons in Fig. \ref{fig:motion_more}, showcasing our InteractMotion against prior methods. In row (a), a cartoon candle is strategically positioned in front of a photorealistic background of person demonstrating blowing action. The composite image accurately portrays the flame changing direction, aligning with the simulated wind direction induced by the action. In (b), exposing a piece of paper to burning flames exhibits signs of burning to the paper, as indicated by the black-ish spots. In (c), ice cream is showing signs of melting when placed on the pan of a kitchen stove, which is correlated with heat. In (d), the candle is extinguished when blown, demonstrating a different outcome from the similar blowing action in (a), highlighting the model's ability to generate diverse responses based on context. The issues of other baselines, noted in the InteractPhys qualitative comparison, are also evident here. Existing methods struggle to preserve the identity of the inserted object—e.g., the candle's appearance is altered in rows (a) and (d). They also fail to account for environmental dynamics like heat distortion in (b–c) or wind effects in (d), resulting in generic composites that ignore background-specific cues. Moreover, blending remains problematic, particularly for Anydoor in (a–b), where transitions appear unnatural. In contrast, our method achieves seamless blending—even across different visual domains, as seen in (a) where a cartoon candle integrates convincingly into a photorealistic human scene.